\documentclass{article}
\usepackage[margin=1in]{geometry}
\usepackage{graphicx}
\usepackage{booktabs}
\usepackage{amsmath}
\usepackage{xcolor}
\usepackage{hyperref}

\title{Running a 20B Video Diffusion Transformer on Consumer Hardware: A Deployment and Measurement Report for MiniMax H3 on Apple Silicon}
\author{Zhenzhou (Luke) Shi \\ \textit{Independent Researcher} \and Yizhe (Steven) Shi \\ \textit{Western Springs College}}
\date{October 2026}

\begin{document}

\maketitle

\textit{Author contributions: L.S. designed the study, performed the deployment and measurements, and wrote the manuscript. S.S. produced the figures and conducted the independent blind review of the production clips (Section 4.7).}

\begin{abstract}
We report the end-to-end deployment and measurement of MiniMax H3, a 20.38-billion-parameter video diffusion transformer (DiT) --- the count measured from the safetensors headers of the Q8 checkpoint as deployed locally (Section 3.2) --- running entirely on consumer Apple Silicon hardware: an Apple Mac mini with M5 Pro chip and 64 GiB of unified memory. Using the official bf16-to-Q8 (8-bit, group size 64) quantization path executed locally on the machine, we produced a working text-to-video pipeline with no cloud dependency. We document the full deployment procedure, including two non-trivial blockers and their fixes, and report measured figures for build time, peak memory, and per-clip render latency across four resolution tiers (Draft 384$\times$640 through Native 768$\times$1344). A 15.08-second 576$\times$1024 clip rendered in 4838.49 seconds with a measured MLX peak of 27.467 GiB. We further operated the system as an unattended production pipeline, completing 17 clips in 26.49 hours with zero render failures. A blind human review of the 17 production clips (generation prompts withheld) averaged 3.12, 3.18, and 2.94 out of 5 for structural integrity, temporal consistency, and visible artifacts. All figures in this report are our own measurements except where explicitly labeled as third-party data. This is a systems/empirical report; we make no claim of model innovation.
\end{abstract}

\section{Introduction}

Cloud video-generation APIs have made high-quality text-to-video broadly accessible, but they carry recurring per-second costs and require uploading prompts --- and often reference assets --- to third-party infrastructure. For sustained, high-volume, or privacy-sensitive iteration, local inference is attractive. The open question we address is a practical one: \textbf{can a 20B-parameter video diffusion transformer run, end to end, on a single consumer machine, and can it do so unattended for hours at a time?}

We answer in the affirmative for one concrete configuration: MiniMax H3, quantized to 8-bit locally, running on an M5 Pro Mac mini with 64 GiB of unified memory via the MLX framework. Our contribution is not a new model or a new quantization method. It is a reproducible deployment record with measured numbers: what breaks, what it costs in time and memory, and what sustained unattended operation looks like on hardware anyone can buy off the shelf.

The intended audience is practitioners and researchers considering local video-model deployment, particularly in resource-constrained or privacy-sensitive settings. All measurements were taken on our own hardware between 11--12 October 2026 (NZDT). We report both successes and blockers; Section 5 discusses the ecosystem gaps this work exposed.

\section{Background}

\subsection{MiniMax H3}

MiniMax H3 (marketed under the Hailuo line) is a video diffusion transformer released by MiniMax under a Community license. The DiT checkpoint we deployed locally contains 20,382,983,040 parameters, counted from the safetensors headers of the five Q8 shards (8-bit weights plus unquantized biases; quantization auxiliary data excluded). We state this as the scale of our deployed artifact, not as a claim about the vendor's canonical model size. The vendor distributes bf16 weights; no official pre-quantized checkpoint was available at the time of our work, so quantization had to be performed locally (see Section 3). We report measurements only; we do not redistribute weights, and we note the Community license constraint in the footnote on the title page.\footnote{MiniMax H3 is distributed under the MiniMax Community license. This report contains measured performance data only. We do not redistribute model weights, and any commercial use or redistribution should be checked against the license terms. Local research use in New Zealand was not blocked by the license.}

\subsection{The MLX / Phosphene inference stack}

Our inference stack is built on Apple's MLX framework, which executes directly on the unified-memory GPU of Apple Silicon. We used Phosphene v4.19.0 together with an H3 runner component. A \texttt{/status} health endpoint returning HTTP 200 confirmed the runner was live with MLX GPU execution enabled. The unified-memory architecture is central to this report: model weights, activations, and the operating system share a single 64 GiB pool, which makes peak-memory measurement --- not just parameter count --- the binding constraint.

\subsection{Quantization: 8-bit, group size 64}

We used the vendor's official \texttt{h3\_build\_q8.sh} script, which converts the bf16 checkpoint to 8-bit quantization with group size 64 (Q8 g64). The official flow is designed for Macs with 36 GiB or more of unified memory and completes the build in approximately 5 minutes on suitable hardware. Quantization is load-bearing here: the full bf16 checkpoint cannot usefully coexist with the build process inside 64 GiB (Section 3.3).

\section{Method}

\subsection{Hardware and software environment}

\begin{itemize}
\item \textbf{Machine:} Apple Mac mini, M5 Pro chip, 64 GiB unified memory, 512 GB internal storage.
\item \textbf{Inference stack:} Phosphene v4.19.0, H3 runner, MLX GPU backend.
\item \textbf{Model storage:} after initial trials, model weights and outputs were migrated to an external volume (\texttt{/Volumes/DevStorage/h3/}); per-file SHA checksums were verified and symlinks left at the original locations. The full Q8 model footprint, including the Q8 pack plus text encoder, VAEs, and configs, measured 55.75 GB. (An earlier figure of approximately 34 GB referred to download size only, not the deployed footprint.)
\end{itemize}

\subsection{Deployment procedure}

\begin{enumerate}
\item Installed Phosphene v4.19.0 and the H3 runner; verified \texttt{/status} returned HTTP 200 with MLX GPU active.
\item Downloaded the bf16 weights (41,399,442,656 bytes, verified 100\% complete) plus the missing components: text encoder, video VAE, audio VAE, and configs.
\item Ran the official \texttt{h3\_build\_q8.sh}, which produced 532 tensors, of which 258 were quantized to 8-bit g64. The build artifact was approximately 20 GB, loaded successfully (\texttt{LOAD OK}), and completed with no out-of-memory event.
\item Resolved two blockers (Section 3.4), then ran smoke tests (Section 4).
\end{enumerate}

\subsection{Memory-boundary methodology}

Because weights, activations, and the OS share one 64 GiB pool, we treated peak MLX-allocated memory as the primary feasibility metric. During the build phase we recorded a memory-gate event (p4): holding bf16 and Q8 resident simultaneously required approximately 58.77 GiB against 64 GiB of physical memory --- uncomfortably close to the ceiling, and the empirical justification for building Q8 and then deleting the bf16 checkpoint (Figure 4). After the build succeeded, the bf16 weights were deleted, leaving 109 GiB of free disk. Peak memory during inference was sampled via MLX's own reporting in each experiment below.

\setcounter{figure}{3}
\begin{figure}[htbp]
\centering
\includegraphics[width=\textwidth]{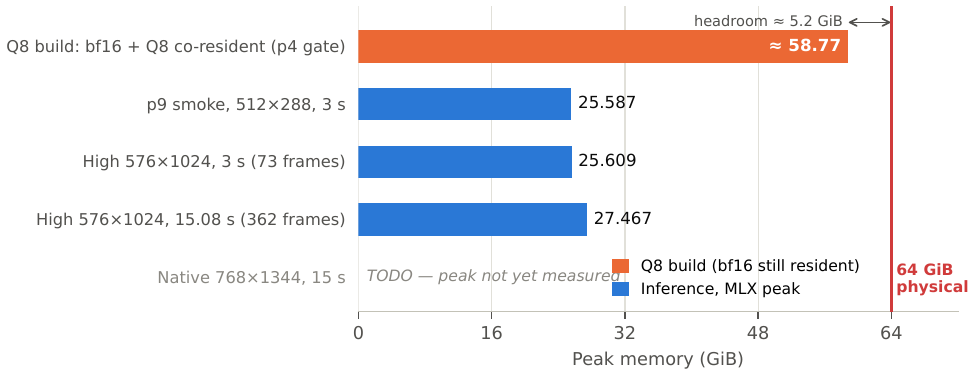}
\caption{Memory budget: peak MLX allocation when holding bf16 and Q8 resident simultaneously against the 64 GiB physical ceiling.}
\end{figure}

\subsection{Blockers and fixes}

\textbf{Blocker 1: hard bf16 dependency in \texttt{h3\_paths()}.} The path-resolution logic assumed bf16 weights would always be present, even in a Q8-only deployment. We resolved this in line with the official flow (download bf16 first, convert locally, then delete), rather than by forking the logic.

\textbf{Blocker 2: text-encoder dependency conflict.} The text encoder required \texttt{fused\_apply} from mlx-vlm v0.6.0, while the installed mlx-vlm was v0.4.4. We applied a \texttt{hasattr}-based compatibility patch and merged it into the runner. This was the only source-level change required for deployment.

\section{Experiments}

All experiments ran on the M5 Pro / 64 GiB machine described in Section 3.1. Outputs were verified with \texttt{ffprobe} and a full decode pass unless otherwise noted. Times are wall-clock.

\subsection{Q8 build}

The official \texttt{h3\_build\_q8.sh} converted the 41,399,442,656-byte bf16 checkpoint into 532 tensors (258 at 8-bit g64), producing an approximately 20 GB artifact that loaded cleanly with no out-of-memory event. The vendor's stated design point --- 36 GiB+ Macs, roughly 5-minute builds --- held on our hardware.

\subsection{First smoke test (p9)}

On 2026-10-11 at 04:07 NZDT, the Q8 pipeline completed its first full text-to-video run: text encoding $\rightarrow$ DiT denoising $\rightarrow$ decoded clip. Parameters: 512$\times$288, 3 seconds, 73 frames, prompt ``a small dog running in a park''. Total wall time was 84.31 seconds with an MLX peak of 25.587 GiB. The output passed \texttt{ffprobe} and full-decode verification.

\subsection{Vertical 9:16 tiers}

The runner natively supports vertical 9:16 output. We tested the High tier (576$\times$1024); the available tiers are Draft 384$\times$640, Standard 448$\times$768, High 576$\times$1024, and Native 768$\times$1344, with 1080p-fit export delivering 1080$\times$1920. A 3-second, 73-frame vertical baseline at High rendered in 640.28 seconds (10 minutes 40 seconds) with an MLX peak of 25.609 GiB, verified by \texttt{ffprobe} and decode (Figure 3).

\setcounter{figure}{2}
\begin{figure}[htbp]
\centering
\includegraphics[width=\textwidth]{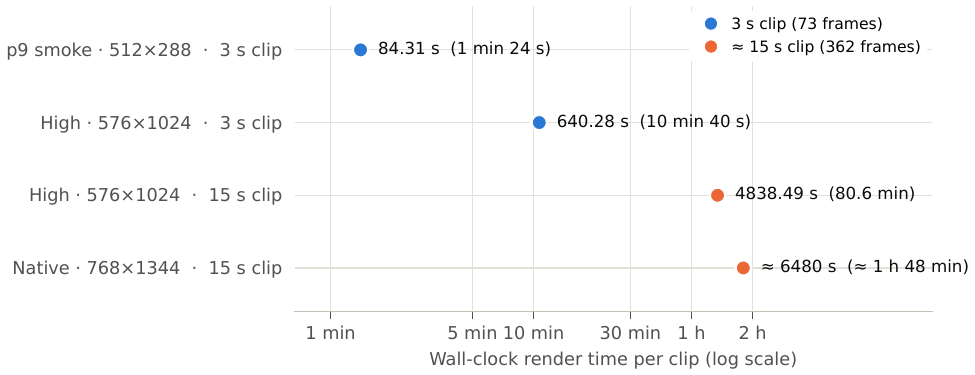}
\caption{Per-clip render time by resolution tier (log scale).}
\end{figure}

\subsection{Full-length vertical clip}

On 2026-10-11 at 11:00 NZDT we rendered a full-length vertical clip: 576$\times$1024 High, 362 frames (approximately 15.08 seconds). Wall time was 4838.49 seconds (approximately 80.6 minutes) with an MLX peak of 27.467 GiB. Output passed \texttt{ffprobe} and full-decode verification.

During a two-hour render window on 2026-10-13/14 we sampled M5 Pro GPU hardware counters with \texttt{powermetrics} at 10-second cadence (720 samples; Figure 9). Mean GPU hardware active residency was 96.8\% (range 3.8\%--100\%) with mean GPU power draw of 31.4 W (max 41.2 W). Utilization stayed near saturation for the full window: on this hardware, H3 rendering is GPU-compute-bound, not memory-bandwidth-bound.

\setcounter{figure}{8}
\begin{figure}[htbp]
\centering
\includegraphics[width=\textwidth]{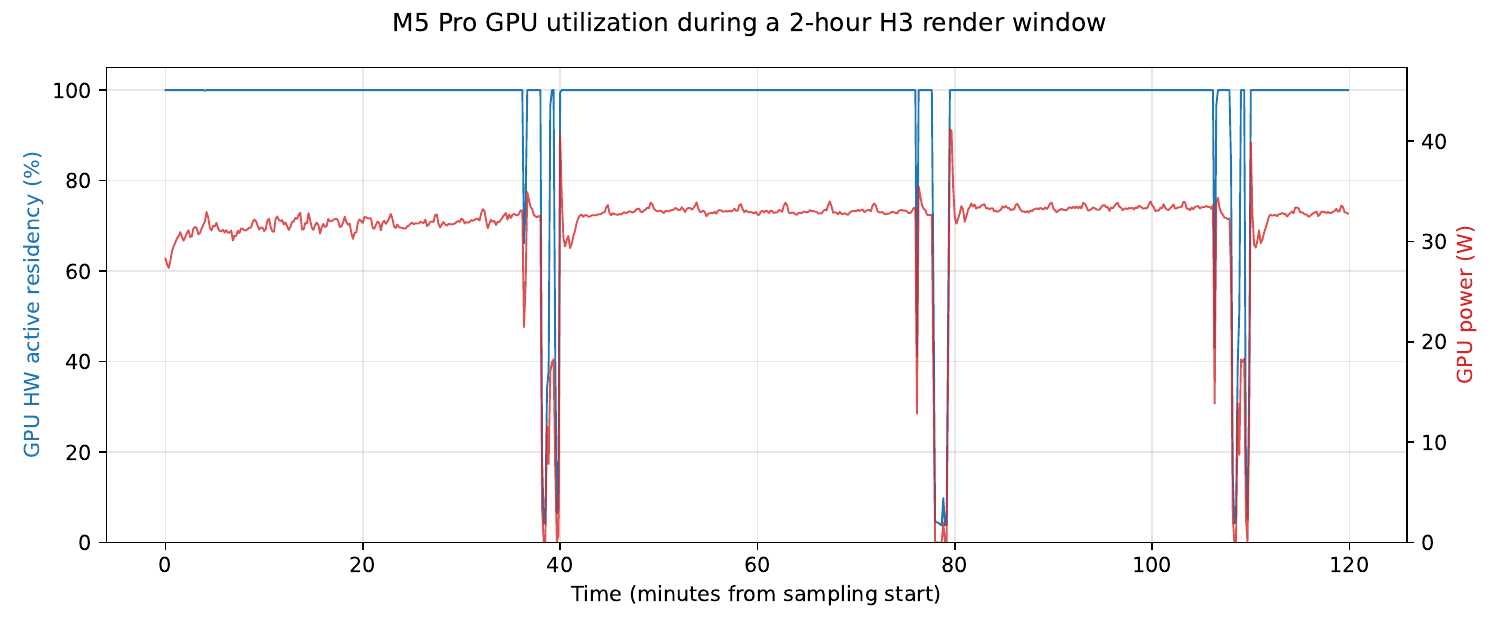}
\caption{M5 Pro GPU hardware active residency and power draw over a two-hour render window (10 s cadence, 720 samples).}
\end{figure}

\subsection{Native tier (768$\times$1344)}

Four consecutive 15-second fashion clips at Native 768$\times$1344 completed with zero failures, each taking approximately 1 hour 48 minutes; this is consistent with an earlier single-clip measurement of 107.49 minutes. The MLX peak-memory figure for the Native tier was recovered from the runner's per-phase metrics records (\texttt{h3\_metrics/*.json}): across 27 completed Native renders, the maximum of each run's per-phase peak (\texttt{max(phases[].peak\_gib)}) is \textbf{29.010 GiB}, reached during the joint-denoise phases. The extraction method was validated against a High-tier record whose computed peak (27.467 GiB) matched the previously reported figure exactly (difference 0.000 GiB).

\subsection{Unattended production pipeline}

The differentiating result of this report is sustained unattended operation. From 2026-10-11 19:37:42 to 2026-10-12 22:07:15 NZDT (26.49 hours), the system operated as a lights-out production pipeline (Figure 1): a scheduler ticked every 60 seconds, dispatched one queued prompt at a time to the renderer, pulled completed clips back, and verified each by hash plus full decode before archiving. In that window it completed 17 clips with \textbf{zero render failures} --- 9 animal-themed (one Standard 448$\times$768, eight High 576$\times$1024) and 8 fashion-themed (all Native 768$\times$1344). Each clip matched exactly one native render record (state done, error null); all 17 archived files were re-verified by full SHA-256 comparison plus complete audio-video decode. Per-clip render times from the renderer's own queue records averaged 1 h 50 min at Native 768$\times$1344 (range 1 h 48 min--1 h 54 min) and 1 h 24 min at High 576$\times$1024 (range 1 h 22 min--1 h 30 min), consistent with the single-clip measurements in Sections 4.4--4.5; the single Standard clip took 25.4 minutes. Inter-completion intervals (Figure 5) bound per-clip time from above, scheduling overhead included, and show the same tier split. Overall throughput was 17 clips in 26.49 hours --- roughly 15 clips per day. (An earlier estimate of 21--22 clips per day is voided; it was based on a lower-quality configuration and we no longer stand behind it.) This is, to our knowledge, the first reported unattended video-DiT production run on consumer Apple Silicon.

\setcounter{figure}{0}
\begin{figure}[htbp]
\centering
\includegraphics[width=\textwidth]{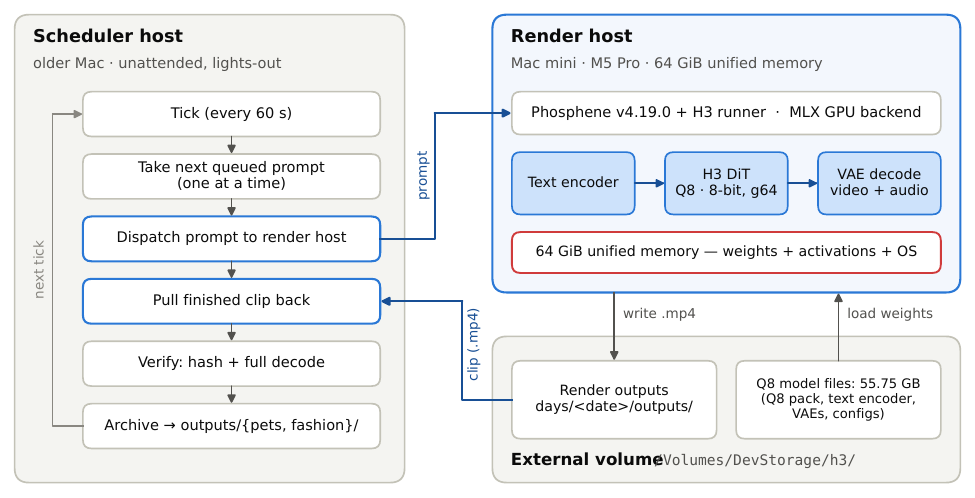}
\caption{System architecture of the unattended H3 production pipeline.}
\end{figure}

\setcounter{figure}{4}
\begin{figure}[htbp]
\centering
\includegraphics[width=\textwidth]{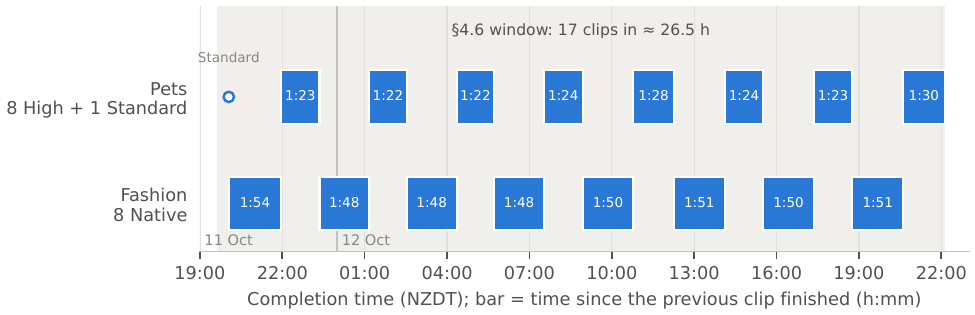}
\caption{Production timeline of the 17-clip unattended run (bars = inter-completion intervals).}
\end{figure}

\subsection{Generation quality: blind human review (our measurements)}

We conducted a blind human review of the 17 production clips from Section 4.6 (362 frames / 15.08 s each). Generation prompts were withheld from the reviewer. Each clip was scored 1--5 (5 = no issue found) on three dimensions: structural integrity (breakage, disappearing objects), temporal consistency (flicker, jitter, abrupt changes), and visible artifacts (smearing, color blocks, illegible text). As a locating aid only, an automated pipeline extracted one frame per second into contact sheets and flagged suspect frames at full resolution; AI-generated draft scores were recorded for comparison but used nowhere in the results.

Clips averaged 3.12 for structure, 3.18 for temporal consistency, and 2.94 for artifacts (Figure 6). Fashion clips scored far lower than pet clips on temporal consistency (2.25 vs 4.00; all nine pet clips scored 4). Artifacts were the weakest dimension overall, with three clips at the minimum score of 1. Recurring failure modes (Figure 8): loss of object permanence within long tracking shots (a frisbee teleporting between the dog's mouth and mid-air), background drift within a single shot, limb and cloth artifacts under fast motion, and illegible pseudo-text. Six of eight fashion clips contain hard cuts, five of them at the same frame (247, $\approx$10.3 s), with lighting, wardrobe, or identity changes across the cut. Sample frames are shown in Figure 7.

Prompt adherence was assessed at unblinding by comparing each blind description against its generation prompt: sixteen clips scored 5/5 and one (fashion-03) scored 4/5, for a mean of 4.94. The frame-247 hard cuts recurred across clips with otherwise high prompt adherence, indicating systematic model behavior rather than prompt-directed editing.

On method: the AI draft scores averaged 1.00 / 0.64 / 1.30 points higher than the human scores (structure / temporal / artifacts), with the largest gaps on fashion clips. One frame per second suffices to locate suspect frames but not to replace human scoring; we report human scores throughout.

\setcounter{figure}{5}
\begin{figure}[htbp]
\centering
\includegraphics[width=\textwidth]{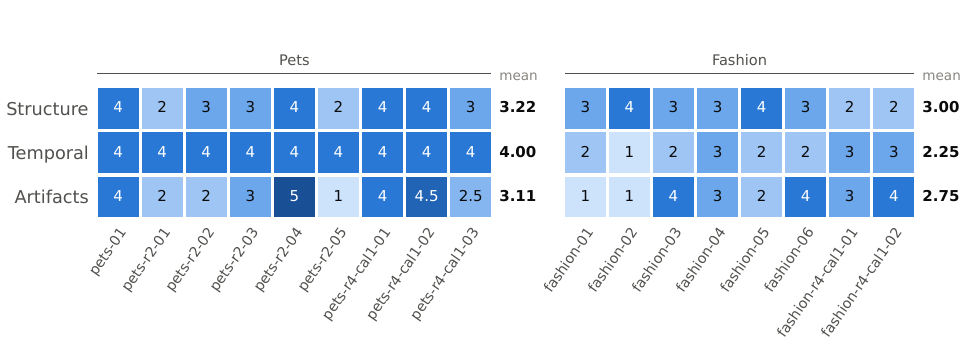}
\caption{Blind human quality scores per clip (heatmap): structural integrity, temporal consistency, visible artifacts.}
\end{figure}

\setcounter{figure}{6}
\begin{figure}[htbp]
\centering
\includegraphics[width=\textwidth]{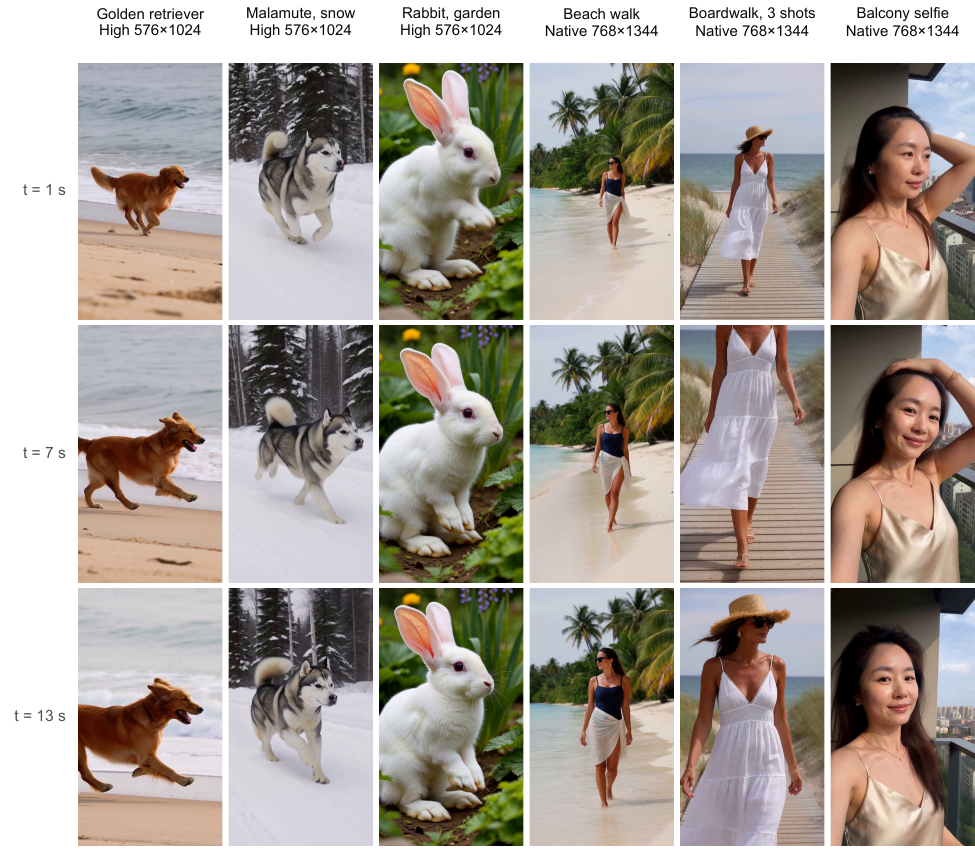}
\caption{Sample frames from 6 clips at 3 timestamps each.}
\end{figure}

\setcounter{figure}{7}
\begin{figure}[htbp]
\centering
\includegraphics[width=\textwidth]{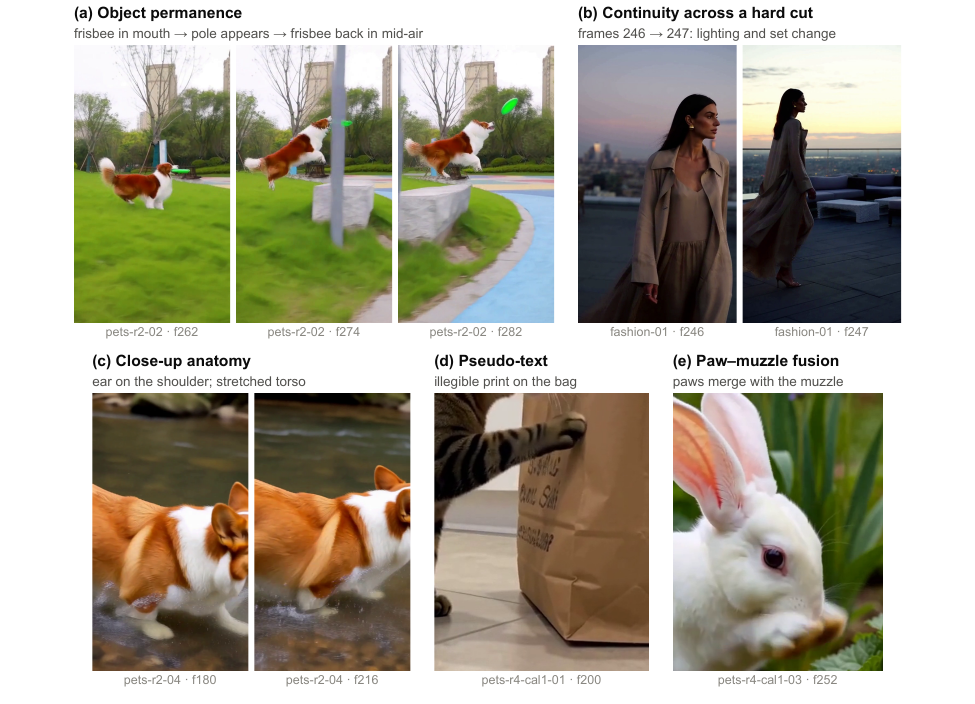}
\caption{Typical failure modes observed in the blind review.}
\end{figure}

\subsection{Quantization fidelity: third-party data (not our measurements)}

We did not run our own bf16-vs-quantized quality comparison (the bf16 checkpoint was deleted after the Q8 build to reclaim disk). We report two third-party sources, clearly labeled as such:

\begin{itemize}
\item \textbf{(a) HuggingFace model page \texttt{pipenetwork/MiniMax-H3-MLX-4bit}:} reports 8-bit vs bf16 PSNR of 27.6 dB with correlation 0.959, described as near-identical; visible artifacts appear at 4-bit and output is described as destroyed at 3-bit. \textit{Third-party data, not our measurement.}
\item \textbf{(b) \texttt{sawfwair/mere-run} benchmark on M4 Max:} reports Q8 vs bf16 SSIM of 0.90--0.92 and PSNR of approximately 29 dB, with no breakdown observed on manual viewing. \textit{Third-party data, not our measurement.}
\end{itemize}

\section{Discussion}

\textbf{The pre-quantized gap.} The most consequential ecosystem finding is mundane: at the time of our work, no official pre-quantized H3 checkpoint existed. Every consumer-hardware user must therefore download the full bf16 weights (41.4 GB in our case) and quantize locally --- a step that itself nearly exhausts a 64 GiB machine (Section 3.3). Pre-quantized community checkpoints (such as the 4-bit MLX page cited above) partially fill this gap, but official Q8 artifacts would remove the most failure-prone step of the deployment.

\textbf{Tooling assumes a ready environment.} Video-generation toolchains in the ComfyUI tradition generally assume the local environment is already correctly configured; our experience is that on Apple Silicon the environment itself --- dependency versions (Section 3.4, Blocker 2), memory ceilings, path assumptions --- is where deployment actually fails. Measurement reports like this one are the missing documentation layer.

\textbf{The 64 GiB gap.} The most comparable published Q8 deployment we found (sawfwair/mere-run [3], M4 Max) runs on 128 GiB hosts, and its catalog explicitly ``makes no 48/64 GB claim.'' Our deployment targets exactly that unclaimed territory: full Q8 text-to-video plus an unattended production pipeline on a 64 GiB consumer machine --- the configuration a buyer can actually take home off the shelf.

\textbf{What the blind review tells us.} The 17-clip review (Section 4.7) moves this report beyond ``it runs'' to ``here is what it gets wrong'': object permanence fails inside long tracking shots, backgrounds drift within a single shot, and a systematic hard cut at frame 247 affects most fashion clips. These are semantic failures, not decode failures --- every clip decoded cleanly --- and they are invisible to the structural verification in Sections 4.2--4.6. For practitioners, the implication is that unattended pipelines need a perceptual review gate, not just a decode gate, before any clip is published.

\textbf{Limitations.} (1) All results are from a single hardware configuration; we make no claim about other Apple Silicon variants beyond noting the M4 Max third-party figures above. (2) Quality scores are from a single human reviewer; inter-rater agreement was not measured. (3) Thermal behavior was not instrumented; GPU power/utilization was sampled during a single two-hour render window (Section 4.4, Figure 9) but not across all tiers.

\section{Conclusion and Future Work}

We demonstrated that a 20B-parameter video diffusion transformer (MiniMax H3) can be deployed, quantized, and operated as an unattended production pipeline on a single consumer machine --- an M5 Pro Mac mini with 64 GiB of unified memory --- with zero cloud dependency. Measured highlights: a 15.08-second 576$\times$1024 clip in 4838.49 seconds at 27.467 GiB MLX peak; 17 consecutive clips in 26.49 hours with zero render failures; and a blind human review averaging 3.12 / 3.18 / 2.94 out of 5 for structural integrity, temporal consistency, and artifacts.

Future work: a same-prompt Q8-vs-bf16 perceptual study if disk budget allows re-downloading bf16; thermal instrumentation across tiers; and evaluating whether the emerging pre-quantized checkpoints reproduce our pipeline results without the local build step.

\end{document}